\def\year{2021}\relax
\documentclass[letterpaper]{article} 
\usepackage{aaai21}  
\usepackage{times}  
\usepackage{helvet} 
\usepackage{courier}  
\usepackage[hyphens]{url}  
\usepackage{graphicx} 
\usepackage{algorithm}
\usepackage{algorithmic}
\usepackage{amsmath}
\usepackage{amsthm,amssymb}
\usepackage{mathrsfs}
\usepackage{natbib}  
\usepackage{caption} 
\usepackage{array}
\usepackage{xcolor}
\usepackage{booktabs}
\usepackage{multirow}

\title{TransTailor: Pruning the Pre-trained Model for Improved Transfer Learning}

\title{TransTailor: Pruning the Pre-trained Model for Improved Transfer Learning}
\author {
        Bingyan Liu\textsuperscript{\rm 1},
        Yifeng Cai\textsuperscript{\rm 2}, 
        Yao Guo\textsuperscript{\rm 1}\thanks{Corresponding author.},
        Xiangqun Chen\textsuperscript{\rm 1} \\
}
\affiliations {
    \textsuperscript{\rm 1}MOE Key Lab of HCST, Dept of Computer Science, School of EECS, Peking University\\
    \textsuperscript{\rm 2}School of Software and Microelectronics, Peking University\\
    \{lby\_cs, caiyifeng, yaoguo, cherry\}@pku.edu.cn
}
\begin{document}
\maketitle

\begin{abstract}
The increasing of pre-trained models has significantly facilitated the performance on limited data tasks with transfer learning. However, progress on transfer learning mainly focuses on optimizing the weights of pre-trained models, which ignores the structure mismatch between the model and the target task. This paper aims to improve the transfer performance from another angle - in addition to tuning the weights, we tune the structure of pre-trained models, in order to better match the target task. To this end, we propose TransTailor, targeting at pruning the pre-trained model for improved transfer learning. Different from traditional pruning pipelines, we prune and fine-tune the pre-trained model according to the \textit{target-aware weight importance}, generating an optimal sub-model tailored for a specific target task. In this way, we transfer a more suitable sub-structure that can be applied during fine-tuning to benefit the final performance. Extensive experiments on multiple pre-trained models and datasets demonstrate that TransTailor outperforms the traditional pruning methods and achieves competitive or even better performance than other state-of-the-art transfer learning methods while using a smaller model. Notably, on the Stanford Dogs dataset, TransTailor can achieve 2.7\% accuracy improvement over other transfer methods with 20\% fewer FLOPs.

\end{abstract}

\section{Introduction}
The remarkable performance achieved by deep Convolutional Neural Networks (CNNs) largely relies on the massive labeled data \cite{krizhevsky2012imagenet}. However, in real-world scenarios where gaining sufficient labeled data through manual labeling is time-consuming and labor-exhausting, we are inevitably confronted with tasks owning only limited labeled data. This situation has motivated the research on \textit{transfer learning} \cite{pan2009survey}, aiming at transferring the knowledge from a related and data sufficient source domain to a resource-constrained (mainly refer to data resource) target domain. 

\begin{figure}[t]
\centering
\includegraphics[width=1\columnwidth]{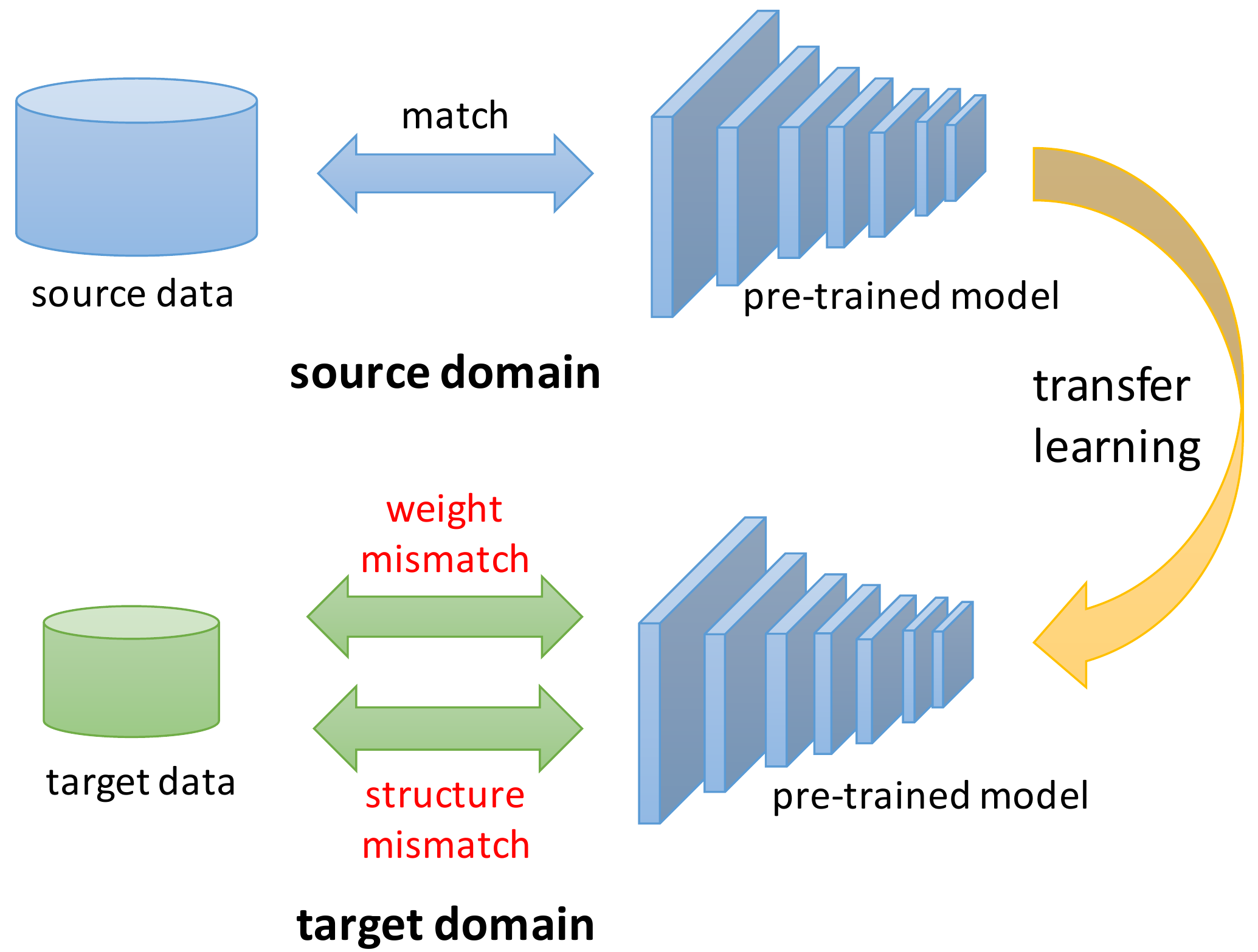} 
\caption{Illustration of the two mismatches during transfer learning.}
\label{fig:transfer-issue}
\end{figure}

Recently, much effort has been made to tune the \textit{pre-trained model} to improve transfer learning. Typically, a pre-trained model refers to a model that has been trained with a large-scale dataset (e.g., ImageNet2012 \cite{deng2009imagenet}) and can serve as a good feature extractor for downstream tasks. A simple yet effective method called \textit{fine-tuning} \cite{yosinski2014transferable} is the most widely used paradigm, which starts with a pre-trained model and further tunes the model weights with the limited target data to achieve transfer learning. In addition, many other regularization methods designed new loss functions to optimize the weights of the pre-trained model, seeking to achieve better knowledge transfer \cite{xuhong2018explicit,li2019delta}.


However, these existing methods may still fail to deliver an optimal solution for transfer learning because they only tune the weights of the pre-trained model. The \textit{model structure}, which is also of vital importance to the final task performance, is ignored since the structure of the pre-trained model is always fixed either in fine-tuning or other transfer schemes. As shown in Figure \ref{fig:transfer-issue}, basically the pre-trained model being transferred may contain two mismatches to the target data: \textit{weight mismatch} and \textit{structure mismatch}. The \textit{weight mismatch} mostly comes from the feature divergence between two tasks and the \textit{structure mismatch} results from the different data scale (i.e., the target task needs a different structure since its data scale is smaller than the source). Previous methods focused on tuning the weights can only tackle the \textit{weight mismatch} issue while ignoring the \textit{structure mismatch} problem.




This paper attempts to improve transfer learning from  a new angle - in addition to tuning the weights, we also tune the structure of the pre-trained model, in order to address the \textit{structure mismatch} problem. Specifically, we apply model pruning techniques \cite{han2015learning} to achieve our goal. Model pruning techniques have been extensively studied in recent years. The key idea is to remove the unimportant weights of a model without incurring much accuracy drop. Adopting pruning techniques is based on the following two insights: \textit{(1) Usually the target task is smaller than the source task and does not need the same model capacity as the pre-trained model provides. (2) Too many parameters may cause overfitting during fine-tuning on the limited target data.} These observations motivate us to prune the pre-trained model to fit the target task. However, directly using traditional pruning algorithms is infeasible because the existing approaches aim at pruning the weights with minimal harm to the original task (source task) rather than the target task. In other words, the weight with high importance to the source task may be insignificant to the target task, given the fact that although there exists some similarity between the two domains, they typically exhibit differences.



To address this issue, we propose TransTailor, targeting at generating an optimal sub-model tailored for a specific target task. Through a learning-based method and a transformation process, TransTailor is able to automatically generate the weight importance for the target data without relying on heuristics, which is then utilized to prune the pre-trained model. After pruning, the computed weight importance is  further incorporated into the fine-tuning process to guide the optimization of weights, in order to better recover the transfer performance of the pruned model. We conduct this pipeline iteratively until we find the optimal sub-model. The major advantage of our approach is that both the pruning and fine-tuning process are based on the target-aware weight importance, which helps to obtain an optimal target-related sub-model. This sub-model can be considered as the significant structure to be transferred in order to improve performance.  Extensive experiments on multiple pre-trained models and datasets demonstrate that TransTailor outperforms the traditional pruning methods and achieves competitive or even better performance compared with other state-of-the-art transfer learning methods while using a smaller model. 




In addition, we would like to highlight that it is not important how much the model can be pruned or accelerated. Instead, the key objective we want to achieve is the task-related optimal sub-net, disregarding its size. In the experiment section, we will analyze the pruned parts of various optimal sub-models generated by TransTailor and summarize some interesting findings. 

This paper makes the following contributions:
\begin{itemize}
\item We introduce the idea of pruning the pre-trained model for improved transfer learning. In addition to tuning the model weights, we also tune the model structure to fit the target task as much as possible. 

\item Based on the idea, we propose TransTailor, in which two techniques, \textit{target-aware pruning} and \textit{importance-aware fine-tuning}, are sequentially executed to respectively prune and fine-tune the pre-trained model according to the properties of the target data.

\item We conduct extensive experiments based on public datasets and model architectures. The evaluation results demonstrate the effectiveness of TransTailor.
    
\end{itemize}

\begin{figure*}[t]
\centering
\includegraphics[width=2.1\columnwidth]{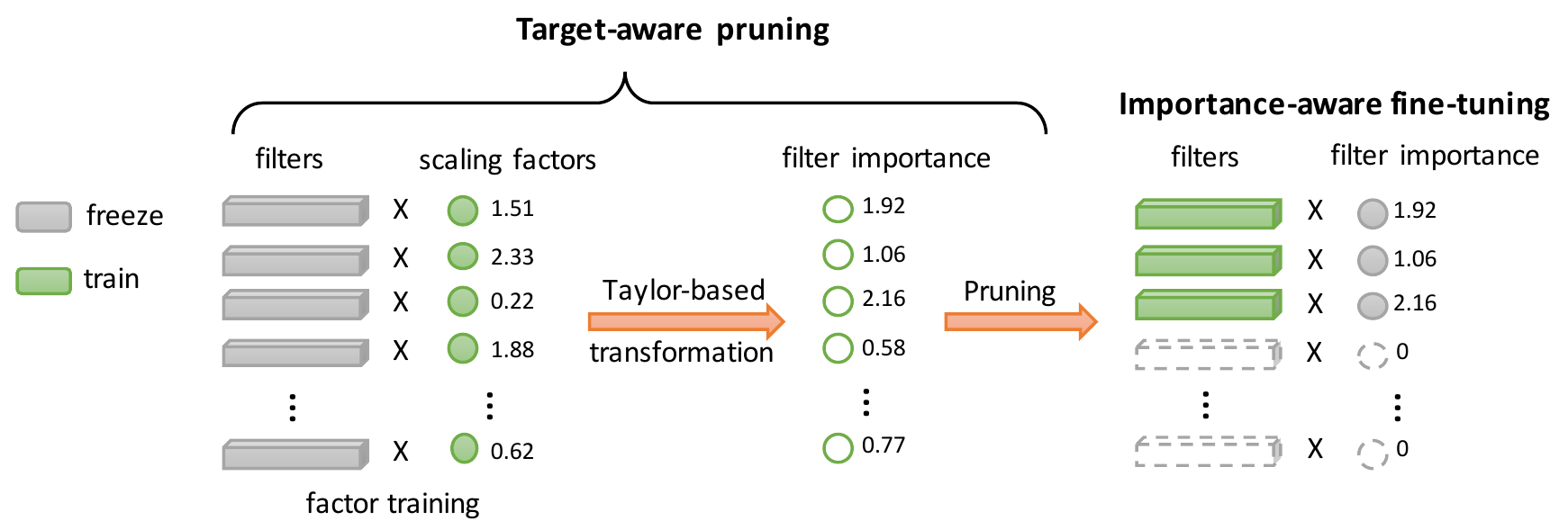} 
\caption{Illustration of the two key techniques of TransTailor. Both \textit{target-aware pruning} and \textit{importance-aware fine-tuning} are based on the target-aware filter importance. Here we use different colors to represent the freezing or training operations.}
\label{fig:overview}
\end{figure*}

\section{Related Work}
\subsection{Transfer Learning}
Transfer learning aims to transfer the information learned from a source task to a target task \cite{pan2009survey}. Our work focuses on inductive transfer learning, where the label space of the target task varies from that of the source task. In this context,  pre-trained models, which are trained with large-scale datasets (e.g., ImageNet2012), have been widely used to conduct fine-tuning or other regularization-based methods. For example, \citeauthor{yosinski2014transferable} \cite{yosinski2014transferable} proposed to freeze the shallow layers and only tune deep layers of the pre-trained model to achieve transfer learning. Xu et al. \cite{xuhong2018explicit} proposed $L^2$-$SP$, which integrates the Euclid distance from the target weights to the starting point as part of the loss, aiming to reduce the distance between source weights and target weights. \citeauthor{li2019delta} \cite{li2019delta} proposed a novel regularization method DELTA to extract the discriminative features from the outputs of the outer layer by a new supervised attention algorithm. Different from these methods that are devoted to tuning weights, we seek to improve transfer learning by tuning both the weight and the structure of the pre-trained model. 



\subsection{Model Pruning}

Model pruning is a promising solution to accelerate CNNs by removing unimportant weights \cite{han2015learning,wen2016learning,lebedev2016fast,liu2019wealthadapt,zhang2020dynamic,liuPMC2020}. Here we focus on filter pruning due to its applicability to any CNN architectures without requiring special software/hardware accelerators. Specifically, \citeauthor{li2016pruning} \cite{li2016pruning} used $\ell_1$-norm to pick unimportant filters and pruned them. \citeauthor{he2018soft} \cite{he2018soft} proposed SFP to enable the pruned filters to be updated when training the model after pruning. \citeauthor{molchanov2016pruning} \cite{molchanov2016pruning} and \citeauthor{you2019gate} \cite{you2019gate} showed the effectiveness of the Taylor expansion method in identifying the global importance of filters. TransTailor also uses filter pruning techniques to tune the structure of the pre-trained model. However, we target at pruning the filters that are less important to the target task rather than the original task as all the existing pruning methods did.


\subsection{Neural Architecture Search}
Neural architecture search (NAS) is dedicated to finding new model architectures in a data-driven way. This is usually achieved using reinforcement learning or evolutionary algorithms to update a model responses to produce architectures with better performance \cite{zoph2016neural,zoph2018learning}. Typically NAS has an extremely large searching space and thus is very time-consuming, which motivates many works on accelerating the search process \cite{liu2018progressive,liu2018darts}. TransTailor can also be seen as an NAS process. However, TransTailor focuses on searching a sub-model from a pre-trained model in order to improve the transfer learning performance, which is substantially different from the traditional NAS.

\section{Method}
\subsection{Overview}
In this paper, we propose TransTailor to tune both the structure and the weight to improve the transfer learning performance. Starting with a pre-trained model, our goal is to generate a sub-model that can best match the target task. To this end, we attempt to design a new pruning method to prune the unimportant weights based on the target task, and a fine-tuning method to  tune the weight for better recovering the transfer performance of the pruned model. Figure \ref{fig:overview} illustrates the two key techniques, which we summarize as follows:
\begin{itemize}
    \item \textit{Target-aware pruning.} We present a new method to learn the weight importance based on the specific target task. Here the importance is measured in a filter-level manner. Specifically, we introduce a scaling factor that is multiplied to the output of each filter and further train them based on the target data. The learned factor values are then transformed based on Taylor expansion, to generate the final filter importance in a global way. Finally, we use the filter pruning technique to prune the model, where unimportant filters can be easily pruned without the need for special software/hardware accelerators for the resulting model.
    
    \item \textit{Importance-aware fine-tuning.}  After the pruning process, each filter of the pruned model still attaches an importance value computed by the previous process. Instead of adopting traditional fine-tuning to recover the accuracy of the pruned model, we incorporate the importance value into the tuning process, in order to provide guidance to the optimization of weights for better transfer performance.
\end{itemize}

We conduct the two techniques iteratively until we find the optimal sub-model.  Note that TransTailor can be generalized to prune any elements of CNNs, such as weight pruning or layer pruning, although here we only focus on filter pruning in this paper. 

\subsection{Problem Definition}
We now introduce the symbols and annotations to formally define the optimization objective. Let $W_f^s$ be the pre-trained model. $f$ is the number of all filters in the pre-trained model and $s$ indicates the model is optimized for source domain. We use $D_s$ and $D_t$ to respectively represent the source data and target data. $\mathcal{L()}$ is the loss function used to optimize the network. Existing filter pruning methods aim at searching a sub-model $W_{f^*}^s$ to minimize the loss increase for $D_s$ under a certain pruning ratio $q\%$, which can be denoted as
\begin{equation}
\begin{split}
    &\arg \min_{W_{f^*}^s} |\mathcal{L}(D_s;W_{f}^s)-\mathcal{L}(D_s;W_{f^*}^s)| \\
    &s.t. \quad f^*=f*q\%
\end{split}
\end{equation}

However, in our transfer learning scenario, we do not care about the loss increase on $D_s$ and are not constrained by the pruning ratio. Instead, we seek to find a sub-model $W_{k}^s$ that can minimize the loss on $D_t$. Here $k$ may take any value as long as it satisfies $k \leq f$. Based on the new scenario, we reformulate the optimization objective as follows
\begin{equation}
    W_o^s=\arg \min_{W_{k}^s} \mathcal{L}(D_t;W_{k}^s)
\end{equation}
where $W_o^s$ represents the optimal sub-model of the pre-trained model. Furthermore, the weights of $W_o^s$ should be tuned with the target data to obtain the final target-aware optimal sub-model $W_o^t$.

In the following sections, we introduce a series of techniques in order to obtain $W_o^t$.

\begin{algorithm}[t]
\caption{The Pipeline of TransTailor}
\label{algorithm}
\begin{flushleft}
    \textbf{Input}: Pre-trained model $W_f^s$, source data $D_s$, target data $D_t$\\
    \textbf{Output}: Optimal sub-model $W_o^t$
\end{flushleft}
\begin{algorithmic}[1] 
\STATE Initialize the scaling factors $\alpha$ randomly
\STATE  Fine-tune the pre-trained model to generate $W_f^{s^*}$
\STATE  Initialize the optimal sub-model $W_o^t= W_f^{s^*}$
\WHILE{(1)}
\STATE Train the factors $\alpha$ by Eq. \ref{train_factor}
\STATE Transform $\alpha$ to the importance vector $\beta$ by Eq. \ref{factor_transform}
\STATE Prune $z$ filters based on $\beta$
\STATE Fine-tune the pruned model by Eq. \ref{train_filter} to generate $W_p^t$ 
\IF{$Acc(D_t;W_o^t) - Acc(D_t;W_p^t)>\tau$}
\STATE break 
\ELSE
\STATE $W_o^t=W_p^t$
\ENDIF
\ENDWHILE
\STATE Get the final optimal sub-model $W_o^t$
\end{algorithmic}
\end{algorithm}

\subsection{Target-aware Pruning}
\subsubsection{Fine-tuning}
\textit{Target-aware pruning} aims at defining the filter importance to the target task. However, the FC (fully-connected) layers of the pre-trained model do not match our target task. For example, the number of neurons in the last FC layer of the ImageNet2012 pre-trained model is 1000 while the target task may not have 1000 classes. Thus we first replace the last FC layer according to the number of 
the target class and then fine-tune it to fit the target task. The fine-tuned pre-trained model is denoted as $W_f^{s*}$. After fine-tuning, the weights in FC layers match the target task and we can focus on estimating the target-aware importance of pre-trained filters. 

\subsubsection{Training the factors}
To measure the filter importance, we introduce a set of tunable scaling factors $\alpha=(\alpha_1^1,\alpha_1^2,...,\alpha_m^n)$, each of which is attached to a pre-trained filter. Here $m$ is the number of layer in the pre-trained model and $n$ is the number of filter in the $m_{th}$ layer. Specifically, for the $j_{th}$ filter of the $i_{th}$ layer $F_i^j$, its output $F_i^j(X)$ is multiplied by the corresponding scaling factor $\alpha_i^j$. Then we train the factors based on the target data to learn a more suitable $\alpha^*$ to minimize the loss
\begin{equation}
\label{train_factor}
    \alpha^*=\arg \min_{\alpha} \mathcal{L}(D_t;W_{f}^s \odot \alpha)
\end{equation}
where $\odot$ represents the element-wise multiplication for each $F_i^j(X)$ and  $\alpha_i^j$. $X$ represents the output of the $(i-1)_{th}$ filter. During the process, pre-trained filters in $W_{f}^s$ are frozen because our goal is to measure the importance of them. 

\subsubsection{Taylor-based transformation}
Although the factor $\alpha^*$ can intuitively represent the filter importance as it is generated in a learning-based method using the target data, there is still a lack of a metric that can define the filter importance with the same scaling factor value while existed in different layers. In other words, it is desirable if we can design a global importance metric over the whole model. Previous works \cite{molchanov2016pruning,you2019gate} have proved the effectiveness of the Taylor expansion method to define the global importance of filters, which can be denoted as
\begin{equation}
    \left|\frac{\partial\mathcal{L}()}{\partial \theta} \theta \right|
\end{equation}
where $\theta$ can be the feature map or a factor related to the corresponding filter. Using this equation, one can calculate the importance of each filter in a global way.

Here we also utilize the Taylor expansion method to transform the learned factor $\alpha^*$ to the vector $\beta$ that represents the global importance. In specific, the importance of $F_i^j$ (i.e., $\beta_i^j$) can be calculated by
\begin{equation}
\label{factor_transform}
    \beta_i^j=\left|\frac{\partial\mathcal{L}(D_t;W_{f}^s \odot \alpha^*)}{\partial (\alpha^*)_i^{j}} (\alpha^*)_i^{j} \right|
\end{equation}

In terms of the vector $\beta$, we can easily compare filters in any location and prune those with low importance values.

\subsection{Importance-aware Fine-tuning}
In the traditional pruning pipeline, fine-tuning is an essential step to recover the accuracy of the pruned model. However, this process treats the whole filters (weights) equally, which may be undesirable as each filter has a unique importance to the target task. Inspired by this, we develop a new fine-tuning scheme based on the target-aware filter importance generated by previous steps. Concretely, we multiply these importance factors to the output of the filter and fix them. Then we only fine-tune the filters using the target data to learn the optimal filter weights to fit the importance as well as minimize the loss
\begin{equation}
\label{train_filter}
    W_{p}^t=\arg \min_{W_{p}^s} \mathcal{L}(D_t;W_{p}^s \odot \beta)
\end{equation}
where the $\beta$ is frozen and $W_{p}^s$ represents the pruned model. After the fine-tuning, we can recover the performance of the pruned model on the target task and obtain $W_{p}^t$. Note that the vector $\beta$ is also integrated into the model to jointly conduct a model inference process.

\begin{table*}[]
\centering
\begin{tabular}{@{}ccccccc@{}}
\toprule
\multicolumn{2}{c}{\multirow{2}{*}{\textbf{Method}}} & \multicolumn{5}{c}{\textbf{Dataset}}                                   \\ \cmidrule(l){3-7} 
\multicolumn{2}{c}{}                                 & Caltech256-30 & Caltech256-60 & CUB-200 & Stanford Dogs & MIT Indoor-67 \\ \midrule
\multirow{2}{*}{FT \cite{tajbakhsh2016convolutional}}                & Top-1 Acc       & 83.7\%        & 85.0\%        & 64.5\% & 89.4\%        & 64.5\%        \\
                                   & FLOPs $\downarrow$       & 0\%           & 0\%           & 0\%    & 0\%           & 0\%           \\ \midrule
\multirow{2}{*}{SFP \cite{he2018soft}}               & Top-1 Acc       & 83.8\%        & 86.9\%        & 78.8\% & 89.0\%        & 77.3\%        \\
                                   & FLOPs $\downarrow$          & 10\%          & 10\%          & 20\%   & 10\%          & 10\%          \\ \midrule
\multirow{2}{*}{GBN \cite{you2019gate}}               & Top-1 Acc       & 79.2\%        & 82.4\%        & 80.7\% & 82.0\%        & 75.2\%        \\
                                   & FLOPs  $\downarrow$         & 10\%          & 10\%          & 30\%   & 10\%          & 20\%          \\ \midrule
\multirow{2}{*}{TransTailor}       & Top-1 Acc       & \textbf{85.3\%}        & \textbf{87.3\%}        & \textbf{80.7\%} & \textbf{91.0\%}        & \textbf{78.2\%}        \\
                                   & FLOPs $\downarrow$          & \textbf{20\%}          & \textbf{30\%}          & \textbf{30\%}   & \textbf{20\%}          & \textbf{20\%}          \\ \bottomrule
\end{tabular}
\caption{Optimal sub-models generated by traditional pruning methods and TransTailor. We mainly compare the accuracy and FLOPs reduction achieved with different schemes.}
\label{tab:pruning_method}
\end{table*}

\begin{table}[]
\centering
\begin{tabular}{@{}ccccc@{}}
\toprule
\multirow{2}{*}{\textbf{Dataset}} & \multicolumn{4}{c}{\textbf{Method}}          \\ \cmidrule(l){2-5} 
                                  & $L^2$     & $L^2$-$SP$   & DELTA  & TransTailor       \\ \midrule
Caltech256-30                     & 84.7\% & 85.4\% & \textbf{85.7\%} & 85.3\% \\
Caltech256-60                     & 87.2\% & 87.2\% & \textbf{87.6\%} & 87.3\% \\
CUB-200                            & 78.4\% & 79.5\% & 78.9\% & \textbf{80.7\%} \\
Stanford Dogs                     & 83.3\% & 88.3\% & 88.3\% & \textbf{91.0\%} \\ \bottomrule
\end{tabular}
\caption{Comparison with state-of-the-art transfer learning methods. TransTailor can achieve comparable or even better performance than other transfer methods with 20\%-30\% FLOPs reduction.}
\label{tab:transfer_method}
\end{table}

\subsection{Searching the Optimal Sub-model}
Based on the aforementioned techniques, we attempt to search out the optimal sub-model $W_o^t$. Specifically, we first initialize the $W_o^t$ with $W_f^{s*}$. Then we iteratively implement our \textit{target-aware pruning} and \textit{importance-aware fine-tuning} techniques. For each iteration, we prune $z$ filters and fine-tune the pruned model, generating a series of sub-models 
\begin{equation}
\label{sub_models}
    W_{f-z}^t, W_{f-2*z}^t, W_{f-3*z}^t, ...
\end{equation}
If the $c_{th}$ sub-model satisfies
\begin{equation}
\label{}
    Acc(D_t;W_o^t) - Acc(D_t;W_{f-c*z}^t)>\tau
\end{equation}
we stop the iteration process, otherwise we use $W_{f-c*z}^t$ to replace the current $W_o^t$.  This logic indicates that the performance of the $c_{th}$ sub-model is significantly worse than the current optimal sub-model, which motivates us to stop the searching and select the current $W_o^t$  as the desirable sub-model. Here the significance degree is controlled by a hyper-parameter $\tau$. An illustration of our overall pipeline is shown in Algorithm \ref{algorithm}.

\section{Experiments}

\subsection{Experimental Settings}
\subsubsection{Pre-trained models and datasets}
We use three pre-trained models with different architectures, including VGG-16 \cite{simonyan2014very}, ResNet-101 \cite{he2016deep} and EfficientNet-B0 \cite{tan2019efficientnet}. All of them are pre-trained with the ImageNet2012 \cite{deng2009imagenet} dataset. We evaluate TransTailor on the following five datasets that are widely used in transfer learning:

\textbf{Caltech 256-30 \& Caltech 256-60} \cite{griffin2007caltech}. Caltech 256 is a dataset of 256 object categories which contains a total of 30607 images. In this paper, we establish two configurations for Caltech 256, with 30 and 60 randomly sampled training examples for each group, in accordance with the procedure used in \cite{xuhong2018explicit}.

\textbf{CUB-200} \cite{wah2011caltech} and \textbf{Stanford Dogs} \cite{khosla2011novel}. CUB-200 contains 11,788 images of 200 species of birds. Stanford Dogs contains photographs of 120 dog classes, each of which contains 100 samples. The two datasets are usually used for fine-grained tasks.

\textbf{MIT Indoor-67} \cite{quattoni2009recognizing}. MIT Indoor-67 is a scene classification dataset of 67 indoor scene categories, each of which consists of 80 training images and 20 test images.

\begin{table}[]
\centering
\begin{tabular}{@{}ccc|cc@{}}
\toprule
\multirow{2}{*}{\textbf{Dataset}} & \multicolumn{4}{c}{\textbf{Method}}          \\ \cmidrule(l){2-5} 
                                  & FT     & FT-Full   &  Ours   
                                  &FLOPs $\downarrow$    \\ \midrule
Caltech256-30                     & 75.9\% & 76.3\% & \textbf{76.4\%} & \textbf{20\%} \\
Caltech256-60                     & 71.4\% & 80.9\% & \textbf{81.8\%} & \textbf{20\%} \\
CUB-200                            & 59.2\% & 78.4\% & \textbf{79.2\%} & \textbf{50\%} \\
Stanford Dogs                     & 81.8\% & 82.6\% & \textbf{84.2\%} & \textbf{20\%} \\ 
MIT Indoor-67                     & 68.3\% & 76.3\% & \textbf{76.5\%} & \textbf{30\%} \\ \bottomrule
\end{tabular}
\caption{Results on VGG-16. We only compare two fine-tuning baselines since few works on transfer learning use this model.}
\label{tab:vgg_result}
\end{table}

\subsubsection{State-of-the-arts}
We compare the proposed approach to two lines of work. The first one is the traditional pruning, which also targets at tuning model structures. Here we implement two state-of-the-art methods: Soft Filter Pruning (SFP) \cite{he2018soft} and Gate Batch Normalization (GBN) \cite{you2019gate} to evaluate the performance on the transfer learning scenario.
The second one is the regularization based transfer learning scheme, which aims to optimize the weights of the pre-trained model. Three regularization approaches, $L^2$, $L^2$-$SP$ \cite{xuhong2018explicit} and DELTA \cite{li2019delta}, are compared to TransTailor. We faithfully record the performance if it is reported in related papers. 

\begin{figure*}[t]
\centering
\includegraphics[width=2.1\columnwidth]{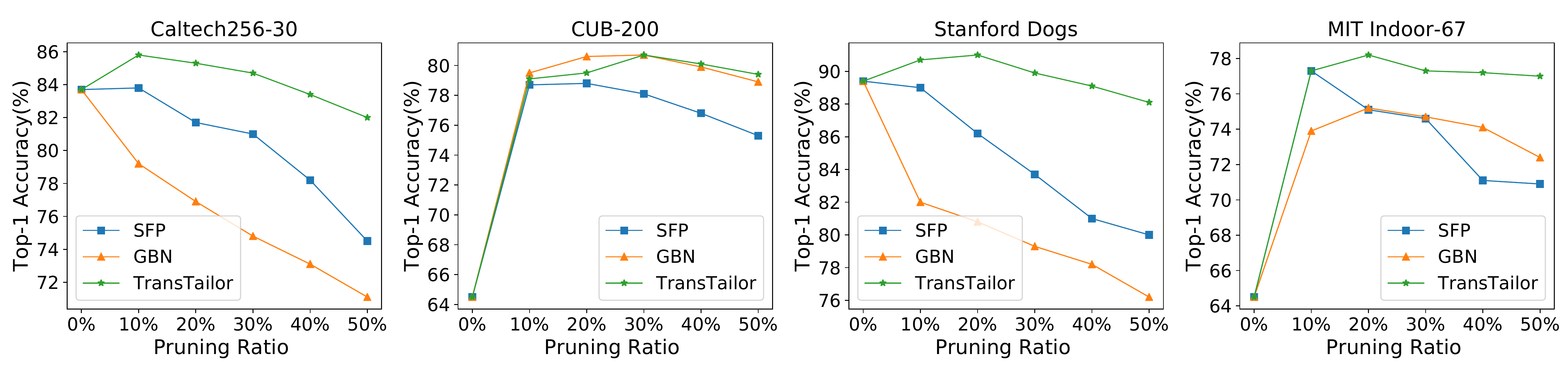} 
\caption{Performance on the generated sub-models during the searching process.}
\label{fig:ablation}
\end{figure*}

\subsubsection{Implementation details}
All experiments are conducted with PyTorch framework. Pre-trained models are provided by Torchvision. The input images are randomly cropped to 224*224 and normalized to zero mean for each channel. All networks are trained using SGD, with 0.005 weight decay and 0.9 momentum respectively. The learning rate is set to 0.005 for the FC layer and 0.0005 for Conv layers. $\tau$ is set to 0.3. After 10$\%$ FLOPs of the pre-trained model is pruned with our \textit{target-aware pruning}, we conduct the \textit{importance-aware fine-tuning} with 40 epochs for ResNet101 and 60 epochs for VGG-16. We iterate this process until the optimal sub-model is selected out. Note that in each iteration, we can flexibly select different pruning budgets (FLOPs reduction) of the pre-trained model and here we use 10$\%$ as an example. All of the experiments are run 3 times and we average them as the reported results. 

\subsection{Ablation Study}
To verify the effectiveness of the proposed \textit{target-aware pruning}, we compare TransTailor with other two pruning methods on ResNet-101, where we mainly focus on the accuracy and FLOPs. Table \ref{tab:pruning_method} summarizes the results. Here FT represents the fine-tuning pipeline that tunes weights of the FC layer based on the target data \cite{tajbakhsh2016convolutional}, which can be considered as a baseline without pruning the structure. Based on the fine-tuned model, we prune its filters with different schemes. The results demonstrate that the Top-1 accuracy achieved by TransTailor outperforms SFP  and GBN  over all datasets as well as reducing more FLOPs, which validates the effectiveness of the proposed approach. Besides, we observe that other pruning methods can also exceed the FT baseline in some cases (e.g., on the CUB-200 dataset), which indicates that the overfitting problem or some ``negative'' pre-trained filters may harm the performance when training on these datasets.

\subsection{Comparison with State-of-the-arts}
Here we conduct comparisons based on the ResNet-101 pre-trained model since most of state-of-the-arts use it as the backbone. Specifically, we compare TransTailor to many transfer learning methods, which improve the transfer performance by only tuning the weights. We record the reported performance and show the results in Table \ref{tab:transfer_method}. Here the comparison on MIT Indoor-67 is omitted because this dataset is based on a different pre-trained model in the related papers. From the results, we can see that TransTailor achieves consistently higher accuracy than $L^2$ and $L^2$-$SP$ baselines over all datasets. In addition, the performance of TransTailor is comparable or even better than DELTA with a smaller model (20\%-30\% FLOPs reduction). Note that our method only utilizes the pruning pipeline, which indicates that if we combine the proposed method with other regularization methods to further tune the weights, greater benefits may be gained.


\subsection{Performance on Sub-models}
During the process of searching the optimal sub-model, a series of sub-models are generated as we state in the previous section. To better observe the trend during the searching process, we implement our method and two other pruning methods step by step, during which we record the accuracy of the pruned model in each iteration. Figure \ref{fig:ablation} demonstrates the results on four datasets. Here we also use  ResNet-101 as the base model.

From the figure, we can draw the following conclusions. First, using our approach, we can always generate a series of pruned models with higher accuracy than the fine-tuned model, which proves that the original pre-trained model does not match the target task. However, SFP and GBN can not well search the optimal sub-model as their pipelines are designed for the source task. Second, when the pruning ratio is large, the superiority of TransTailor is more significant. For example, if we prune 50\% FLOPs of the pre-trained model, TransTailor can exceed other baselines by 8.1\% on the Stanford Dogs dataset, which further suggests the importance of the target-aware pruning if we focus on acceleration.

\subsection{Results on Other Model Architectures}
Besides the complex ResNet-101, we also evaluate the performance on VGG-16 and EfficientNet-B0, in order to further validate the effectiveness of TransTailor. Since there are few works of transfer learning based on these models, we only compare our method to the FT baseline and its variant FT-Full (i.e., fine-tuning all the parameters). Table \ref{tab:vgg_result} summarizes the results on VGG-16. We can clearly see that TransTailor yields consistently better results than the two baselines while reducing 20\%-50\% FLOPs. Besides, we use the EfficientNet-B0 to evaluate our approach on Stanford Dogs and find that we can outperform FT and FT-Full by 7.0\% and 0.7\% with 20\% fewer FLOPs. This demonstrates the effectiveness of our pipeline for lighter models.


\begin{figure*}[]
\centering
\includegraphics[width=2\columnwidth]{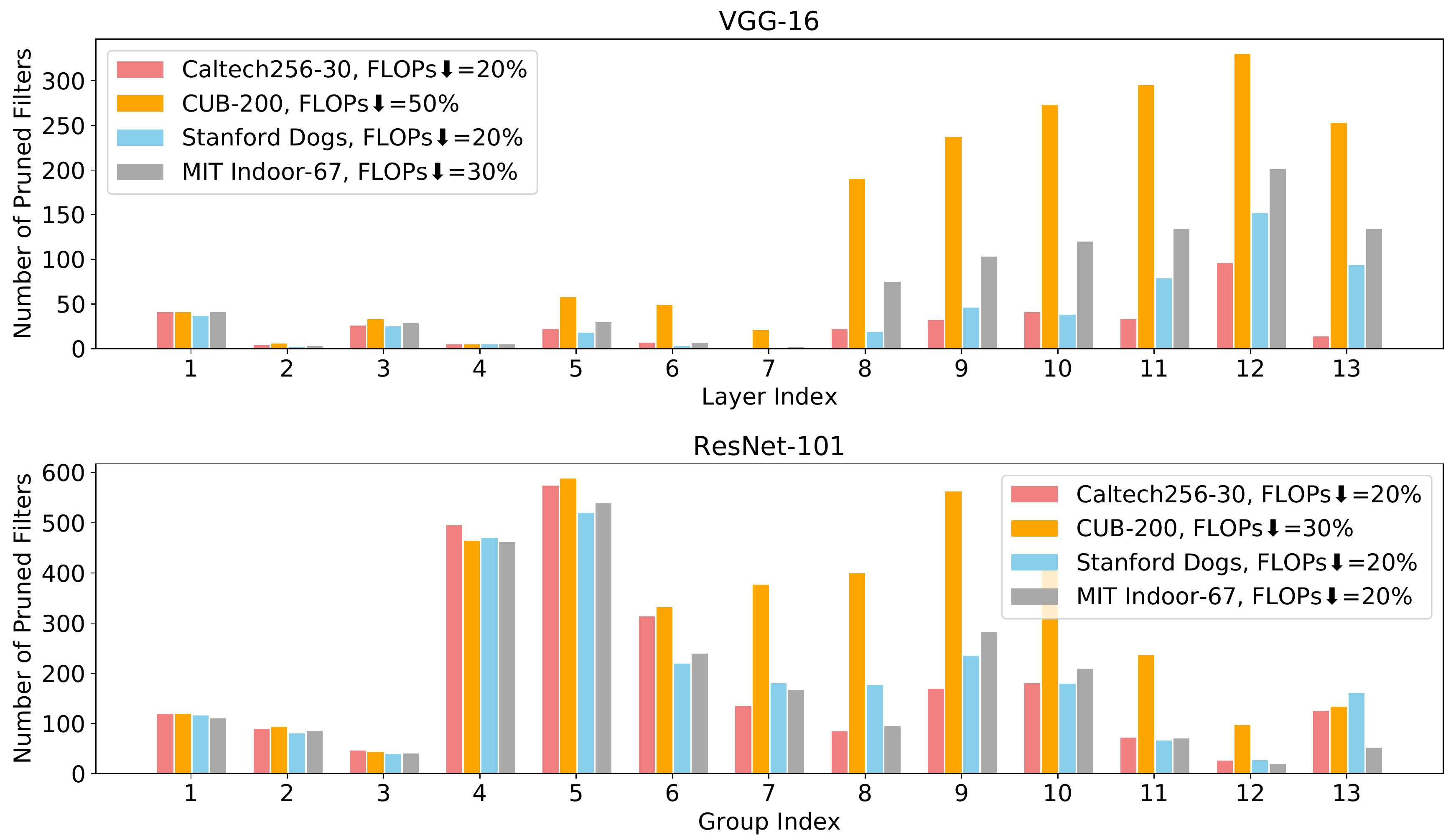} 
\caption{Distribution of the pruned filters on different models and datasets. The FLOPs$\downarrow$ represents the ratio of reduced FLOPs on the generated optimal sub-model.}
\label{fig:pruned_filter}
\end{figure*}

\subsection{Distribution of Pruned Filters}
To observe the optimal sub-models generated by TransTailor, we visualize the distribution of the pruned filters on VGG-16 and ResNet-101. Specifically, we respectively select different datasets and record the corresponding pruned filters in each Conv layer.

\subsubsection{Pruned filters on VGG-16}
We show the optimal sub-models tailored for 4 different datasets on the top of Figure \ref{fig:pruned_filter}. We can clearly see that the pruned filters of these sub-models differ from each other, indicating that TransTailor is a task-driven structure tuning algorithm. Concretely, two phenomena can be found: (1) Generally the number of pruned filters in shallow layers is significantly less than deep layers. This demonstrates that the shallow layers of VGG-16 hold more low-level general image features, which should be reserved due to their generalization to any image recognition tasks. (2) For the last Conv layer (layer index=13), TransTailor tends to prune fewer filters compared with other deep layers. We believe this is because the last Conv layer is more related to the final classification, which may have a more complex feature process and thus have little redundancy.


\subsubsection{Pruned filters on ResNet-101}
For ResNet-101, it is hard to display the pruned filters in each Conv layer since there are too many Conv layers in the network. Thus, we partition them into 13 groups in order to make a fair comparison to VGG-16. We illustrate the filter distribution of the optimal sub-models on the bottom of Figure \ref{fig:pruned_filter}. From the figure, we can make the following observations. First, although the size of most sub-models is identical (FLOPs reduction=20\%), their filter distribution in each layer is still diverse. For example, in the $13th$ group, TransTailor can prune 125 and 161 filters for Caltech256-30 and Standford Dogs. However, on MIT Indoor-67, our method can only prune 52 filters despite having the same FLOPs reduction. Second, we surprisingly find that the pruned filters do not have the same regularity as we observe in VGG-16 (i.e., shallow layers prune less, deep layers prune more). Instead, it seems no specific regularity. Based on a recent work, which shows that ResNets behave as ensembles of shallow classifiers \cite{veit2016residual}, we believe this phenomenon is reasonable because the ensemble effect may diminish the assumption that shallow layers should be shared with common low-level features. 


\section{Conclusion}
This paper improves transfer learning from a new angle: by tuning both the model structure and the model weights. We realize this goal through TransTailor, an approach aiming at generating the target best structure by pruning techniques. We design the pruning metric and the fine-tuning scheme in terms of the \textit{target-aware} filter importance, which can be measured by a learning-based method and a transformation process. Extensive experiments confirm the effectiveness of the proposed approach. In the future, we will attempt to combine the structure tuning and existing regularization schemes to further improve transfer learning.

\section{Acknowledgments}
We would like to thank the anonymous reviewers for their valuable feedback.    This work was partly supported by the National Key Research and Development Program (2016YFB1000105) and the National Natural Science Foundation of China (61772042).

\bibliography{TransTailor}

\begin{thebibliography}{29}
\providecommand{\natexlab}[1]{#1}
\providecommand{\url}[1]{\texttt{#1}}
\providecommand{\urlprefix}{URL }
\expandafter\ifx\csname urlstyle\endcsname\relax
  \providecommand{\doi}[1]{doi:\discretionary{}{}{}#1}\else
  \providecommand{\doi}{doi:\discretionary{}{}{}\begingroup
  \urlstyle{rm}\Url}\fi

\bibitem[{Deng et~al.(2009)Deng, Dong, Socher, Li, Li, and
  Fei-Fei}]{deng2009imagenet}
Deng, J.; Dong, W.; Socher, R.; Li, L.-J.; Li, K.; and Fei-Fei, L. 2009.
\newblock Imagenet: A large-scale hierarchical image database.
\newblock In \emph{2009 IEEE conference on computer vision and pattern
  recognition}, 248--255. Ieee.

\bibitem[{Griffin, Holub, and Perona(2007)}]{griffin2007caltech}
Griffin, G.; Holub, A.; and Perona, P. 2007.
\newblock Caltech-256 object category dataset.
\newblock \emph{California Institute of Technology} .

\bibitem[{Han et~al.(2015)Han, Pool, Tran, and Dally}]{han2015learning}
Han, S.; Pool, J.; Tran, J.; and Dally, W. 2015.
\newblock Learning both weights and connections for efficient neural network.
\newblock In \emph{Advances in neural information processing systems},
  1135--1143.

\bibitem[{He et~al.(2016)He, Zhang, Ren, and Sun}]{he2016deep}
He, K.; Zhang, X.; Ren, S.; and Sun, J. 2016.
\newblock Deep residual learning for image recognition.
\newblock In \emph{Proceedings of the IEEE conference on computer vision and
  pattern recognition}, 770--778.

\bibitem[{He et~al.(2018)He, Kang, Dong, Fu, and Yang}]{he2018soft}
He, Y.; Kang, G.; Dong, X.; Fu, Y.; and Yang, Y. 2018.
\newblock Soft filter pruning for accelerating deep convolutional neural
  networks.
\newblock \emph{Proceedings of the International Joint Conference on Artificial
  Intelligence (IJCAI)} .

\bibitem[{Khosla et~al.(2011)Khosla, Jayadevaprakash, Yao, and
  Li}]{khosla2011novel}
Khosla, A.; Jayadevaprakash, N.; Yao, B.; and Li, F.-F. 2011.
\newblock Novel dataset for fine-grained image categorization: Stanford dogs.
\newblock In \emph{Proc. CVPR Workshop on Fine-Grained Visual Categorization
  (FGVC)}, volume~2.

\bibitem[{Krizhevsky, Sutskever, and Hinton(2012)}]{krizhevsky2012imagenet}
Krizhevsky, A.; Sutskever, I.; and Hinton, G.~E. 2012.
\newblock Imagenet classification with deep convolutional neural networks.
\newblock In \emph{Advances in neural information processing systems},
  1097--1105.

\bibitem[{Lebedev and Lempitsky(2016)}]{lebedev2016fast}
Lebedev, V.; and Lempitsky, V. 2016.
\newblock Fast convnets using group-wise brain damage.
\newblock In \emph{Proceedings of the IEEE Conference on Computer Vision and
  Pattern Recognition}, 2554--2564.

\bibitem[{Li et~al.(2016)Li, Kadav, Durdanovic, Samet, and
  Graf}]{li2016pruning}
Li, H.; Kadav, A.; Durdanovic, I.; Samet, H.; and Graf, H.~P. 2016.
\newblock Pruning filters for efficient convnets.
\newblock \emph{arXiv preprint arXiv:1608.08710} .

\bibitem[{Li, Grandvalet, and Davoine(2018)}]{xuhong2018explicit}
Li, X.; Grandvalet, Y.; and Davoine, F. 2018.
\newblock Explicit inductive bias for transfer learning with convolutional
  networks.
\newblock In \emph{International Conference on Machine Learning}, 2825--2834.

\bibitem[{Li et~al.(2019)Li, Xiong, Wang, Rao, Liu, and Huan}]{li2019delta}
Li, X.; Xiong, H.; Wang, H.; Rao, Y.; Liu, L.; and Huan, J. 2019.
\newblock Delta: Deep learning transfer using feature map with attention for
  convolutional networks.
\newblock \emph{International Conference on Learning Representations (ICLR)} .

\bibitem[{Liu, Guo, and Chen(2019)}]{liu2019wealthadapt}
Liu, B.; Guo, Y.; and Chen, X. 2019.
\newblock WealthAdapt: A General Network Adaptation Framework for Small Data
  Tasks.
\newblock In \emph{Proceedings of the 27th ACM International Conference on
  Multimedia}, 2179--2187.

\bibitem[{Liu et~al.(2020)Liu, Li, Liu, Guo, and Chen}]{liuPMC2020}
Liu, B.; Li, Y.; Liu, Y.; Guo, Y.; and Chen, X. 2020.
\newblock PMC: A Privacy-preserving Deep Learning Model Customization Framework
  for Edge Computing.
\newblock \emph{Proceedings of the ACM on Interactive, Mobile, Wearable and
  Ubiquitous Technologies} 4(4).

\bibitem[{Liu et~al.(2018)Liu, Zoph, Neumann, Shlens, Hua, Li, Fei-Fei, Yuille,
  Huang, and Murphy}]{liu2018progressive}
Liu, C.; Zoph, B.; Neumann, M.; Shlens, J.; Hua, W.; Li, L.-J.; Fei-Fei, L.;
  Yuille, A.; Huang, J.; and Murphy, K. 2018.
\newblock Progressive neural architecture search.
\newblock In \emph{Proceedings of the European Conference on Computer Vision
  (ECCV)}, 19--34.

\bibitem[{Liu, Simonyan, and Yang(2018)}]{liu2018darts}
Liu, H.; Simonyan, K.; and Yang, Y. 2018.
\newblock Darts: Differentiable architecture search.
\newblock \emph{arXiv preprint arXiv:1806.09055} .

\bibitem[{Molchanov et~al.(2017)Molchanov, Tyree, Karras, Aila, and
  Kautz}]{molchanov2016pruning}
Molchanov, P.; Tyree, S.; Karras, T.; Aila, T.; and Kautz, J. 2017.
\newblock Pruning convolutional neural networks for resource efficient
  inference.
\newblock \emph{International Conference on Learning Representations (ICLR)} .

\bibitem[{Pan and Yang(2009)}]{pan2009survey}
Pan, S.~J.; and Yang, Q. 2009.
\newblock A survey on transfer learning.
\newblock \emph{IEEE Transactions on knowledge and data engineering} 22(10):
  1345--1359.

\bibitem[{Quattoni and Torralba(2009)}]{quattoni2009recognizing}
Quattoni, A.; and Torralba, A. 2009.
\newblock Recognizing indoor scenes.
\newblock In \emph{2009 IEEE Conference on Computer Vision and Pattern
  Recognition}, 413--420. IEEE.

\bibitem[{Simonyan and Zisserman(2014)}]{simonyan2014very}
Simonyan, K.; and Zisserman, A. 2014.
\newblock Very deep convolutional networks for large-scale image recognition.
\newblock \emph{arXiv preprint arXiv:1409.1556} .

\bibitem[{Tajbakhsh et~al.(2016)Tajbakhsh, Shin, Gurudu, Hurst, Kendall,
  Gotway, and Liang}]{tajbakhsh2016convolutional}
Tajbakhsh, N.; Shin, J.~Y.; Gurudu, S.~R.; Hurst, R.~T.; Kendall, C.~B.;
  Gotway, M.~B.; and Liang, J. 2016.
\newblock Convolutional neural networks for medical image analysis: Full
  training or fine tuning?
\newblock \emph{IEEE transactions on medical imaging} 35(5): 1299--1312.

\bibitem[{Tan and Le(2019)}]{tan2019efficientnet}
Tan, M.; and Le, Q.~V. 2019.
\newblock Efficientnet: Rethinking model scaling for convolutional neural
  networks.
\newblock \emph{arXiv preprint arXiv:1905.11946} .

\bibitem[{Veit, Wilber, and Belongie(2016)}]{veit2016residual}
Veit, A.; Wilber, M.~J.; and Belongie, S. 2016.
\newblock Residual networks behave like ensembles of relatively shallow
  networks.
\newblock In \emph{Advances in neural information processing systems},
  550--558.

\bibitem[{Wah et~al.(2011)Wah, Branson, Welinder, Perona, and
  Belongie}]{wah2011caltech}
Wah, C.; Branson, S.; Welinder, P.; Perona, P.; and Belongie, S. 2011.
\newblock The caltech-ucsd birds-200-2011 dataset.
\newblock \emph{California Institute of Technology} .

\bibitem[{Wen et~al.(2016)Wen, Wu, Wang, Chen, and Li}]{wen2016learning}
Wen, W.; Wu, C.; Wang, Y.; Chen, Y.; and Li, H. 2016.
\newblock Learning structured sparsity in deep neural networks.
\newblock In \emph{Advances in Neural Information Processing Systems},
  2074--2082.

\bibitem[{Yosinski et~al.(2014)Yosinski, Clune, Bengio, and
  Lipson}]{yosinski2014transferable}
Yosinski, J.; Clune, J.; Bengio, Y.; and Lipson, H. 2014.
\newblock How transferable are features in deep neural networks?
\newblock In \emph{Advances in neural information processing systems},
  3320--3328.

\bibitem[{You et~al.(2019)You, Yan, Ye, Ma, and Wang}]{you2019gate}
You, Z.; Yan, K.; Ye, J.; Ma, M.; and Wang, P. 2019.
\newblock Gate decorator: Global filter pruning method for accelerating deep
  convolutional neural networks.
\newblock In \emph{Advances in Neural Information Processing Systems},
  2133--2144.

\bibitem[{Zhang et~al.(2020)Zhang, Li, Guo, Chen, and Liu}]{zhang2020dynamic}
Zhang, Z.; Li, Y.; Guo, Y.; Chen, X.; and Liu, Y. 2020.
\newblock Dynamic slicing for deep neural networks.
\newblock In \emph{Proceedings of the 28th ACM Joint Meeting on European
  Software Engineering Conference and Symposium on the Foundations of Software
  Engineering}, 838--850.

\bibitem[{Zoph and Le(2016)}]{zoph2016neural}
Zoph, B.; and Le, Q.~V. 2016.
\newblock Neural architecture search with reinforcement learning.
\newblock \emph{arXiv preprint arXiv:1611.01578} .

\bibitem[{Zoph et~al.(2018)Zoph, Vasudevan, Shlens, and Le}]{zoph2018learning}
Zoph, B.; Vasudevan, V.; Shlens, J.; and Le, Q.~V. 2018.
\newblock Learning transferable architectures for scalable image recognition.
\newblock In \emph{Proceedings of the IEEE conference on computer vision and
  pattern recognition}, 8697--8710.

\end{thebibliography}

\end{document}